\documentclass{article}

\PassOptionsToPackage{numbers, compress}{natbib}

\usepackage[preprint]{neurips_2020}

\usepackage{graphicx}
\usepackage{stfloats}
\usepackage{pdflscape}
\usepackage{listings}

\usepackage{amsmath,amsfonts,bm}
\usepackage{wrapfig}
\usepackage{caption}
\usepackage{subcaption}
\usepackage{algorithm}
\usepackage{algorithmic}

\usepackage[utf8]{inputenc} 
\usepackage[T1]{fontenc}    
\usepackage{hyperref}       
\usepackage{url}            
\usepackage{booktabs}       
\usepackage{amsfonts}       
\usepackage{nicefrac}       
\usepackage{microtype}      

\title{Sparse Meta Networks for Sequential Adaptation and its Application to Adaptive Language Modelling}

\author{%
  Tsendsuren Munkhdalai \\
  Microsoft Research\\
  Montr\'{e}al, Qu\'{e}bec, Canada\\
  \texttt{tsendsuren.munkhdalai@microsoft.com} \\
}

\begin{document}

\maketitle

\begin{abstract}
    Training a deep neural network requires a large amount of single-task data and involves a long time-consuming optimization phase. This is not scalable to complex, realistic environments with new unexpected changes. 
    Humans can perform fast incremental learning on the fly and memory systems in the brain play a critical role.
    We introduce Sparse Meta Networks -- a meta-learning approach to learn online sequential adaptation algorithms for deep neural networks, by using deep neural networks. 
    We augment a deep neural network with a layer-specific fast-weight memory. The fast-weights are generated sparsely at each time step and accumulated incrementally through time providing a useful inductive bias for online continual adaptation. We demonstrate strong performance on a variety of sequential adaptation scenarios, from a simple online reinforcement learning to a large scale adaptive language modelling.
    
\end{abstract}

\section{Introduction}

    Deep neural networks are typically trained on a large amount of a single task data through a time-consuming optimization phase. This assumes that the distribution over data points is fixed. However, such neural models do not scale to complex, realistic environments and are prone to distributional shifts and adversarial data points~\cite{szegedy2013intriguing}. Online learning on the other hand does not make any distributional assumption and naturally involves an adversarial scenario. However, due to the larger number of training parameters and non-convex optimization landscape, the deep neural networks are hard to train in online settings~\cite{robbins1951stochastic,dauphin2014identifying}.
    
    \begin{wrapfigure}{r}{0.6\textwidth}
      \begin{center}
        \includegraphics[width=0.55\textwidth]{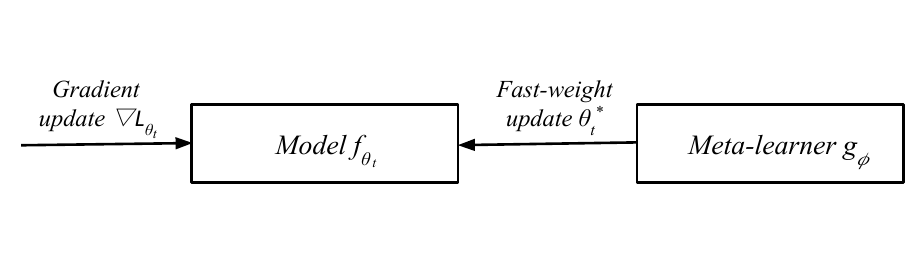}
        \caption{\small Schematic illustration of the update procedure in MetaNet. In the MetaNet framework, a neural net denoted as $f_{\theta_t}$ is exposed to two types of update: the loss gradient and the fast-weights updates. The fast-weights are generated by a meta-learner network based on the gradients w.r.t the slow-weights. The fast-weights maintain information at a time scale that is longer than the network activations, but shorter than the slow-weights, enabling a fast adaptation.
        }
        \label{fig:metanet_update}
     \end{center}
     \vskip -0.45in
    \end{wrapfigure}
    
    Meta-learning (i.e. learning-to-learn)~\cite{schmidhuber1987,mitchell1993explanation,andrychowicz2016learning} has emerged as a promising technique for fast training of deep neural networks by acquiring and transferring knowledge across different tasks through a learned learning algorithm. This work proposes a meta-learning approach to learn sequential adaptation algorithms for deep neural networks. We introduce a sparse variant of Meta Networks~\cite{pmlr-v70-munkhdalai17a} to perform an online and continual fast adaptation of deep neural networks over a data stream with non-stationary distribution. 
    
    In Sparse Meta Networks (Sparse-MetaNet), fast-weights~\cite{schmidhuber1992learning,hinton1987using} are generated sparsely at each step by a meta-learner and accumulated across multiple steps. When the sparse fast-weights are accumulated in this way, across different tasks, they all together act as an mixture of multiple experts in a single Sparse-MetaNet model. Such sparsely generated recurrent fast-weights are not only computationally efficient; and thus can be applied with large scale deep neural networks, but also crucial to maintain a far past memory over the streaming data. 
    
    To demonstrate the effectiveness of our approach, we introduce a new vision based benchmark called Online Cifar. In the Online Cifar setup, Sparse-MetaNet shows a better flexibility and a less catastrophic interference, and achieves the best classification accuracy compared with gradient based baselines. We also evaluate Sparse-MetaNet on Wisconsin Card Sorting Test (WCST), a simple online reinforcement learning problem adapted from the human cognitive test~\cite{berg1948simple} and large scale language modelling benchmarks. When used along with Transformer-XL~\cite{dai2019transformer} for adaptive language modelling, our Sparse-MetaNet achieves 1.00 bpc on enwik8 and 22.67 perplexity on WikiText-103 datasets, improving upon the original Transformer-XL result of 1.06 bpc and 24.0 perplexity, respectively. 
    
    \begin{wrapfigure}{r}{0.5\textwidth}
      \begin{center}
        \includegraphics[width=0.50\textwidth]{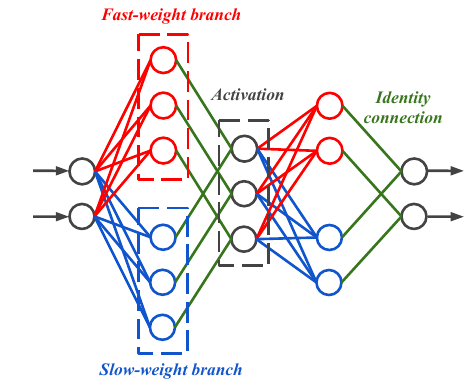}
        \caption{\small A two-layer feed-forward MetaNet with 2 inputs, 2 outputs and 3 hidden units. The red and blue lines indicate the fast and slow-weight connections. The green lines are identity connection. The shared slow-weight branches generalize across tasks whereas the fast-weight branches specialize for each task and both are optimized jointly in a MetaNet model.
        }
        \label{fig:ff_metanet}
     \end{center}
     \vskip -0.45in
    \end{wrapfigure}
    
\section{Method}
    We focus on the standard online learning setup where the data is sequentially made available over time. The model receives a new input $x_t$ and makes a prediction $\hat{y_t}$ at time step $t$. The model is then provided with the true label $y_t$ and now it must update its internal state in order to improve the prediction performance for the following time steps. The distribution of the incoming data is not fixed and shifts over time due to some stimuli. The model has no knowledge about when the distributional shift and the stimuli occur. So it has to infer this only from the data and must adapt to the distribution shift quickly to minimize error. In this work, we explicitly train a deep neural network for such a fast and continual online adaptation. The core component of our approach is the Meta Networks with fast-weight memory.
    
\subsection{Meta Networks}
    The Meta Networks~\cite{pmlr-v70-munkhdalai17a} were originally introduced in the context of few-shot learning. In  Figure~\ref{fig:metanet_update}, we illustrated the overall update procedure in MetaNet. In the MetaNet framework, we perform two types of model updates.  Like the standard neural networks, we update the weight parameters with the loss gradient by using an optimization algorithm. However, such a gradient based weight is not fast enough due the inherent bias of the iterative gradient optimizers~\cite{robbins1951stochastic}. Therefore, the model receives its second update based on fast-weight parameters. 
    The fast-weight update enables a quick reaction of the model in response to external stimuli whereas the gradient update on the standard slow-weights requires more data and time to be effective. 
    
    The fast-weights maintain a long-term memory and MetaNet can be understood as a memory augmented neural network (MANN)~\cite{graves2014neural,weston2014memory}. Unlike the conventional MANNs, MetaNets can augment each layer of an arbitrary neural architecture with a fast-weight memory to promote layer-wise fast adaptations. Also, the fast-weights are generated based on the loss gradient. We use a meta-learner module -- another artificial neural network to generate and optimize the fast-weights. 
    
    A two-layer feed-forward neural net with the MetaNet architecture is shown in Figure~\ref{fig:ff_metanet}. The network has fast and slow-weight branches for each layer. The slow-weight branch is shared for all tasks in a similar way as multi-task learning~\cite{Caruana93multitasklearning} while the fast-weight branch is specialized for each input and task.
    The outputs from the two branches are integrated with the identity connection~\cite{he2016deep} for a layer activation. The forward computation is then given as follows\footnote{The original MetaNet setup applies non-linearity $\sigma$ before element-wise summation. We found that this sometimes outputs a zero vector for layer activation when used with ReLU.}:
    \begin{equation}
        \label{eq:metanet_forward}
        h^{l}_t =  \sigma (W^{l}_t h^{l-1}_{t} + M^{l}_{t} h^{l-1}_{t} + b^{l})
    \end{equation}
    where $\sigma$ is the non-linearity and $l$ indexes the layer of the network. $W^{l}_t$ and $M^{l}_{t}$ are the slow and fast-weights at layer $l$. 
    
    The fast-weights are generated by a meta-learner module $g_\phi$ based on the loss derivative w.r.t to the slow-weights ${W^{l}_{t-1}}$ at time step $t-1$ as: 
    
    \begin{equation}
    \label{eq:fw_generation}
        M^{l}_{t} = g_\phi(\nabla_{W^{l}_{t-1}}\mathcal{L}_t)
    \end{equation}
    where $\mathcal{L}_t$ denotes the incurred loss. 
    
    In the original MetaNet architecture for few-shot learning, the parameterization of the meta-learner $g_\phi$ involved a slot-based memory~\cite{marblestone2020product} and attention mechanism~\cite{bahdanau2014neural}. A set of fast-weights was first generated for all inputs with target labels and stored in memory slots. The memory slots are then queried on the fly by using the attention mechanism to generate the fast-weights for each new input. This enabled the every input to assemble its fast-weight transformation, effectively building a new model for each input. 
    
    Although such example-level fast-weights provide a fine-grained adaptation and flexibility to the model for few-shot learning, it is computationally expensive in the case of the continuous adaptation with a long sequential data stream.
    Moreover, it becomes exceedingly difficult to use the fast-weights to adapt a neural networks with large number of parameters, preventing its application to large scale problems, i.e. Transformer-based language modelling.
    
    \begin{wrapfigure}{r}{0.6\textwidth}
    \vskip -0.3in
    \begin{minipage}[t]{0.6\textwidth}
    \begin{algorithm}[H]
    \caption{Sparse-MetaNet for Sequential Adaptation}
    \label{alg:sparse_metanet}
    \begin{algorithmic}[1]
    {\footnotesize
    \STATE choose hyper-parameters $\beta_1$, $\beta_2$, $\gamma$ and $p$
    \STATE choose BPTT length $k$
    \STATE randomly initialize slow-weights $\theta=\lbrace W \rbrace$ and meta-weights $\phi=\lbrace W_{meta} \rbrace$
    \STATE initialize fast-weights $M_{0} \leftarrow 0$ and gradient average $I_{0} \leftarrow 0$
    
    \FOR{$t = 1,\dots$} 
        \STATE Receive new input batch $\lbrace x_t \rbrace$
        \STATE Predict targets $\lbrace \hat{y}_{t} \rbrace$ as in Eq.~\ref{eq:metanet_forward} and shown in Figure~\ref{fig:ff_metanet}
        \STATE Receive true target $\lbrace y_t \rbrace$
        \STATE Incur loss $\mathcal{L}_t$ and compute its gradient $\nabla_{\theta} \mathcal{L}_t$ and $\nabla_{\phi} \mathcal{L}_t$
        \STATE Update gradient average $I_{t} \leftarrow \gamma I_{t-1} + \beta_{1} \nabla_{\theta} \mathcal{L}_t$
        \IF { $k$ \textbf{divides} $t$ }
            \STATE Update slow-weights $\theta$ and meta-weights $\phi$ using $\nabla_{\theta} \mathcal{L}_t$ and $\nabla_{\phi} \mathcal{L}_t$ with a SGD optimizer
            \STATE Truncate computation graph
            \STATE $M_{t} \leftarrow M_{t-1}$
        \ELSE
            \STATE Sample binary indicator mask $A_{t}$ from ${\rm \it Bernoulli}(p)$
            \STATE Compute sparse fast-weights $M_{sparse,t}$ as in Eq.~\ref{eq:sparse_metalearner}
            \STATE Update fast-weights $M_{t} \leftarrow (1 - A_{t}) M_{t-1} + M_{sparse,t}$
        \ENDIF
    \ENDFOR
    }
    \end{algorithmic}
    \end{algorithm}
    \end{minipage}
    \end{wrapfigure}
    
\subsection{Sparse Meta Networks for Sequential Adaptation}
    In the Sparse Meta Networks (Sparse-MetaNet), we introduce several crucial and efficient modifications to the MetaNet approach for fast sequential adaptation with deep neural networks. 
    
    Inspired by Nesterov’s accelerated gradients~\cite{nesterov1983method,sutskever2013importance,dozat2016incorporating}, we make use of the past gradients in the meta-learner. A moving average of the past gradients is defined as 
    \begin{equation}
    \label{eq:metainfo}
        I_{t} \leftarrow \gamma I_{t-1} + \beta_{1} \nabla_ {W^{l}_{t}} \mathcal{L}_t
    \end{equation}
    and then the fast-weights are calculated as $M^{l}_{t} = g_\phi(I_{t} + \beta_{2} \nabla_{W^{l}_{t}} \mathcal{L}_t)$ where $\gamma$, $\beta_1$ and $\beta_2$ are hyper-parameters that can be meta-trained or tuned using a validation data. One can view $I_{t}$ as a compressive memory that tracks the relation between the model input and expected output via the gradient averaging.
   
    The meta-learner can then be simplified into a single feed-forward neural net without any attention and slot-based memory components for efficiency. The meta-learner performs the following forward pass computation for each fast-weight element specified by indices $i$ and $j$: 
    \begin{equation}
    \label{eq:metalearner}
        M^{l}_{i,j,t} \leftarrow W^{3}_{Meta} \sigma(W^{2}_{Meta} \sigma(W^{1}_{Meta} (I_{i,j,t} + \beta_{2} \nabla_{W^{l}_{i,j,t}} \mathcal{L}_t))).
    \end{equation}
   The meta-learner operates coordinate-wise and computes the each fast-weight element in parallel. The weights of the meta-learner are shared for the elements of the fast-weights.
    
    Now we introduce a sparse recurrent process in the fast-weight computation for efficient sequential adaptation with deep neural networks. In Sparse-MetaNet, at each time step, a sparse fast-weight matrix $M^{l}_{sparse,t}$ is generated. We first sample a binary indicator mask matrix $A^{l}_{t}$ with the same size as $M^{l}_{sparse,t}$. Each element of this matrix is set to 1 or 0 with probability $p$ or $1-p$, respectively. Then a new fast-weight element $M^{l}_{i,j,sparse,t}$ is obtained only if its corresponding mask entry $A^{l}_{i,j,t}$ is equal to 1, otherwise simply set to 0:
    \begin{equation}
    M^{l}_{i,j,sparse,t} \leftarrow \begin{cases}
        \label{eq:sparse_metalearner}
        g_\phi(I_{i,j,t} + \beta_{2} \nabla_{W^{l}_{i,j,t}} \mathcal{L}_t) & A^{l}_{i,j,t} = 1 \\
        0 & A^{l}_{i,j,t} \neq 1
    \end{cases}
    \end{equation}
    
    Therefore, in Sparse-MetaNet, the meta-learner computes the Eq.~\ref{eq:metalearner} for only few elements at a single time step when a small probability $p$ is used. Since we need not even construct the computation graph for the most of the fast-weight elements, Sparse-MetaNet is extremely efficient and scales to large scale neural networks. In practice, we choose a small probability value for parameter $p$ (i.e. $p=0.05$) during training to speed-up and then during test, we increase the parameter or select a suitable value based on a validation data for efficiency.
    
    Once the sparse-fast weights are generated, we recursively compute cumulative fast-weights $M^{l}_{t}$ over time as
    \begin{equation}
    \label{eq:sparse_fw}
        M^{l}_{t} \leftarrow (1 - A^{l}_{t}) M^{l}_{t-1} + M^{l}_{sparse,t}
    \end{equation}
    where $A^{l}_{t}$ is the binary indicator mask with each element drawn independently from $A^{l}_{i,j,t} \sim {\rm \it Bernoulli}(p)$.
    The recurrent fast-weights $M^{l}_{t}$ maintain a mixture of sparse fast-weights when accumulated over multiple time steps, and this sparse and recurrent process provides a useful inductive biases for an incremental learning.
    
    In Algorithm~\ref{alg:sparse_metanet}, we summarize the training procedure for Sparse-MetaNet models. The overall training algorithm is similar to the truncated back-propagation through time~\cite{werbos1990backpropagation} in learning recurrent neural networks (RNNs) and is a sequential extension of the original MetaNet training algorithm. The algorithm introduces only single additional hyper-parameter $k$ to specify the length of the back-propagation through time (BPTT). 
    
    During training, we alternate between the fast-weight and the gradient updates. We generate and apply the fast-weights for prediction for $k-1$ subsequent steps. After that, for every $k$-th step, we jointly update the slow-weight parameters of the main network and the meta-weights of the meta-leaner by using a SGD optimizer. Like RNNs, the computation graph is tracked for $k$ steps and then truncated and like RNN recurrent states, temporal information are gathered in the fast-weights over time and made available to be readily exploited in the following time steps. 
    
    Our algorithm scales linearly with input sequence length, and together with a Sparse-MetaNet model, it provides a bounded memory learning in a continuously changing environment. Although, we have focused mainly on feed-forward neural network for clarity, our approach is generic enough to be used with other network types such as Convolutional networks and Transformers as shown in the experiments.

\section{Related Work}
    There is a very broad line of work on online learning with convex functions~\cite{cesa2006prediction,freund1999adaptive,hoi2018online}. One of the earliest online learning methods is the Perceptron algorithm~\cite{rosenblatt1958perceptron}. Other more sophisticated methods are kernel Perceptron~\cite{aiserman1964theoretical}, Winnow~\cite{littlestone1988learning}, second order Perceptron~\cite{cesa2005second} and online gradient descent~\cite{bottou1998online,zinkevich2003online}. The online gradient descent algorithm can readily be applied to deep neural networks to learn from a streaming data. However, multiple iterative passes over each data point is normally required due to the non-convex optimization landscape~\cite{robbins1951stochastic,dauphin2014identifying} and the training of deep neural networks with the online gradient updates is usually difficult~\cite{sahoo2017online}.
    
    Recurrent neural networks with an external memory  system~\cite{hochreiter1997long,munkhdalai2016neural,munkhdalai2019metalearned,schlag2018learning} exhibit a  sequential adaptation capability when the data stream is provided in a custom format~\cite{schmidhuber1992learning,santoro2016meta,mishra2017simple}. The present work is along the similar line with an advantage that our Sparse-MetaNet approach is generic and can be applied to an arbitrary neural network architecture. More recently, \cite{nagabandi2018deep} explored a similar setup as ours with a focus on online reinforcement learning in changing environments~\cite{al2017continuous}. Their approach involves maintaining a mixture of neural network models over the entire data stream and this can be quite expensive to scale to large neural networks. A more complex online meta-learning scenario was also introduced~\cite{finn2019online} based on the MAML algorithm~\cite{finn2017model} where the meta-learner is continually trained while we in this work focus on meta-learning a learned online learning algorithm that can generalize on unseen test data stream.
    
    More broadly, our approach falls into the category of memory-based meta-learning~\cite{ortega2019meta}. Meta-learning~\cite{schmidhuber1987,mitchell1993explanation,andrychowicz2016learning} has been extensively studied in few-shot learning setup~\cite{koch2015siamese,vinyals2016matching,bansal2019learning} where a meta-learning algorithm assumes an access to  a distribution of tasks with a few labelled examples each. 
    Unlike online learning, task identities are known and the examples are not strictly provided in an sequential order in few-shot learning. Moreover, the online learning setup does not restrict the number of examples to be only a few. Therefore, a learned online learning algorithm should be capable of a fast adaptation when there is only a few examples available from a new distribution, but also scalable to a larger data regime to be able to continually improve its underlying model. The later can be challenging for learned gradient-based learning algorithms~\cite{andrychowicz2016learning,finn2017model} due to short horizon biases introduced during training~\cite{wu2018understanding}. Specially, the learned learning rate, an important hyper-parameter for gradient-based training does not generalize across different data regimes~\cite{antoniou2018train}.
    
    Continual learning~\cite{kirkpatrick2017overcoming,aljundi2019continual,vuorio2018meta,aljundi2019online} investigates the catastrophic interference~\cite{mccloskey1989catastrophic} in neural networks in a similar setup as online learning. However, the prior continual learning algorithms have mainly been focused on a small handful set of tasks~\cite{pmlr-v80-serra18a}. The Online Cifar benchmark introduced in this work offers an interesting testbed to evaluate continual learning approaches on a large task stream.

\section{Experiments}
In this section, we first investigate the efficiency of our proposed approach on two toy tasks which provide well-controlled testbeds. Our models are compared with fine-tuning and online SGD baselines as those baselines are simple to implement and yet have strong performance~\cite{dhillon2019baseline}. We then focus on more practical applications of the proposed method. Particularly, we apply Sparse-MetaNet to Transformer-based large scale language modelling.
\subsection{Toy Tasks}
\subsubsection{Wisconsin Card Sorting Test}
    Wisconsin Card Sorting Test (WCST)~\cite{berg1948simple} is a cognitive reasoning and flexibility test for humans. A participant is asked to classify a selected card into one of 4 different categories and is told whether the classification is correct or not. The classification can be made according to one of 3 different rules. It can be based on the color, the shape, or the number of symbols on each card. The classification rule changes over time and the participant is not told when the rule has changed nor the rule itself. So in order to accurately classify the cards, the participant has to figure out the rule by a sequence of trial-and-errors. The test measures how well one can adapt to the changes. Specially, we are interested in how quickly one can figure out a new rule with less perseveration error. The perseveration error is the amount of the misclassifications made by applying the old rule to the cards corresponding to the new rule.
    
    We trained neural network agents in reinforcement learning setup. The rules for the tasks can be represented by three different optimal policy. In this online setup, the RL agent has to learn to quickly switch between those task policies. Each episode consists of a sequence of 16 cards and if the agent categorize total 48 cards over 3 subsequent episodes without any error, we consider that the current task is solved and the rule is found. We then switch to a new random rule in one of the next 50 episodes. This ensures that the agent has no access to the schedules of the rule switching and new task. The code snippet to generate WCST tasks are given in Appendix.
    
    The neural agents have 2 fully connected layers with 256 hidden units and ReLU activations, followed by the action and value heads and are trained with the Advantage Actor-Critic (A2C) algorithm~\cite{mnih2016asynchronous}. For the Sparse-MetaNet agent, we augmented the all layers including the action and value heads with the fast-weights. The slow and meta-weights were optimized using Adam. The the mask probability $p$ was set to 0.3. We update the slow and meta-weights after every 2 fast-weight updates  by using the BPTT length $k=3$.
    
    For baselines, we have two offline agents and their online variants with Adam~\cite{kingma2014adam} and RMSProp~\cite{hinton2012neural} optimizers. The offline agents are trained from scratch for each task with new rule and do not transfer any information across the tasks whereas in the case of Sparse-MetaNet and the online methods, we continue to train a single agent across the tasks. Therefore, the offline agents have access to the task schedules (i.e. changes of the rules) whereas the Sparse-MetaNet and online agents do not. The learning rates for Adam and RMSProp were selected by a grid search over $\lbrace 0.0001, 0.0003, 0.0005, 0.001, 0.003 \rbrace$.
    
    We found out that the standard online agents fail in this RL setting due to a forward interference and negative transfer. Specially, training of the agents diverged after around 10 tasks and the agents failed to converge later on. We showed the results of the offline agents in separate plots for clarity in Appendix.
    
\begin{figure*}[!t]
        \begin{center}
        \includegraphics[width=1\textwidth]{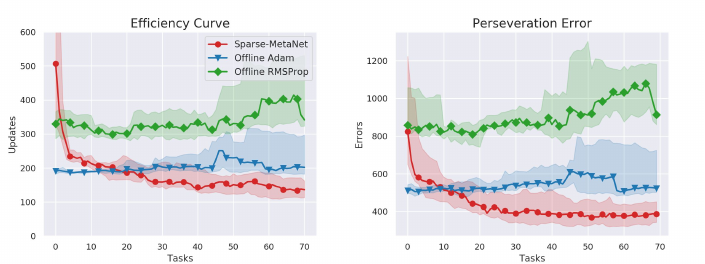}
        \caption{Wisconsin Card Sorting Test results. Left: the number of updates required to solve each task. The total number for Sparse-MetaNet includes both gradient and fast-weight updates. Right: total perseveration errors made by each agent before switching to the next task.}
        \label{fig:wcst}
        \end{center}
        \vskip -0.2in
    \end{figure*}

    The final results (averaged over 10 runs) of the Sparse-MetaNet and offline agents are shown in Figure~\ref{fig:wcst}. Our Sparse-MetaNet agent shows sample efficiency by performing a forward transfer across the WCST tasks. The effectiveness of Sparse-MetaNet is more clear in terms of the perseveration error. The Sparse-MetaNet agent demonstrates more flexibility by making less perseveration errors after being exposed to only 10 tasks.

\subsubsection{Online Cifar}
    Online Cifar is an online 5-way classification benchmark built from the Cifar-100 dataset. In this benchmark, an algorithm receives a sequence of image and class label pairs from a task $\tau_n$. The task sequence then continues for $T_{\tau_n}$ rounds before changing to the next task with different distribution. The model has no access to the task length $T_{\tau_n}$ and the lengths vary across the tasks.
    
    For each new task, we create new task-level labels by ensuring the following 3 criteria: 
    \begin{itemize}
        \item Subsequent tasks $\tau_{n-1}$ and $\tau_n$ share examples with the same original Cifar label, but different task-level class labels. We use this set to measure the perseveration error. \item Each task $\tau_n$ must include examples that have the same global Cifar and local task-level labels as the previous task $\tau_{n-1}$. This set is used to measure incremental learning ability and catastrophic interference in neural models while acquiring novel concepts introduced in the current task. \item The third criteria ensures the novel input and label pairs to be included in the current task $\tau_{n-1}$. 
    \end{itemize} 
    For each task, we allocate examples of 1 class label for the perseveration error and another 2 class labels for the interference error estimations. The remaining 2 class labels were used to introduce novel concepts. We provided code snippets for generating such tasks in Appendix~\footnote{Our train/dev/test splits can be downloaded at <link-will-be-available>}.
    
    On this benchmark, an efficient online  sequential adaptation algorithm should achieve a higher cumulative accuracy with lower perseveration and interference error rates over the input task stream. Moreover, we split the original Cifar data into 40/30/30 classes for train/dev/test sets. So the models are evaluated on unseen classes during test. This allows us to further evaluate the generalization capability of learned online learning algorithms.  
    
    We designed the following baseline methods based on pretraining and online fine-tuning. \\
    \textbf{Online fine-tuning}: In an online fine-tuning baseline, we train the model on the sequence of training tasks and evaluate by performing online gradient updates on the test data. We reset the optimizer's state only at the beginning of the entire test stream. \\
    \textbf{Online fine-tuning with pretraining}: We have another baseline where the model is pretrained on the full training classes for multiple epochs and then fine-tuned on the test stream in the same online fashion as the previous baseline. \\
    \textbf{Fine-tuning with pretraining}: The third baseline is also based pretraining but this method assumes an access to the task length. For each task, the neural net is initialized back to its pretrained weights and the optimizer states are reset. This baseline is similar to the standard fine-tuning practices with the following exception. 
    
    For each baseline, we perform an early stopping based on the validation performance obtained with the same procedure as the test one. Therefore, rather than a fixed evaluation, the models are validated based on its online adaptation performance on the held-out validation data for early stopping.
    
    \begin{figure*}[!t]
        \begin{center}
        \includegraphics[width=1\textwidth]{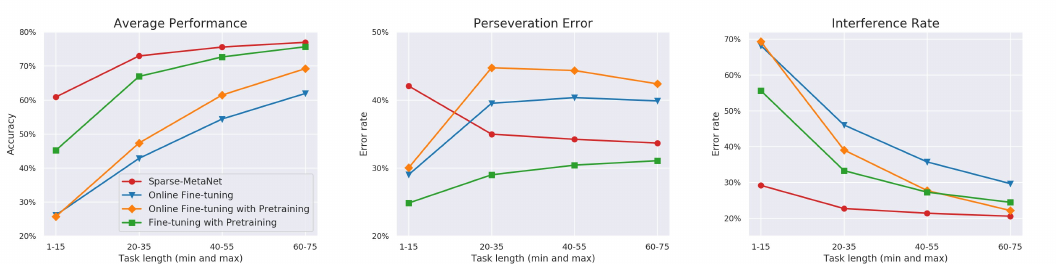}
        \caption{Online Cifar test results. Left: average task accuracy over each test stream. Center: perseveration error rate and lower is better. Right: catastrophic interference measured at two subsequent tasks. }
        \label{fig:online_cifar}
        \end{center}
        \vskip -0.2in
    \end{figure*}
    
    We used a 12-layer ResNet~\cite{he2016deep} with each of these baselines as well as our Sparse-MetaNet. For the Sparse-MetaNet model, all 12 layers except the last softmax layer were augmented with the fast-weights. We set the BPTT length $k$ to 3 and the mask probability $p$ to 0.3.  During test, we increased the mask probability to 0.5 based the validation result and only updated the fast-weights for a fair comparison. Specially, we did not perform the gradient-based update, although this can further boost the result.
    
    The training data stream consisted of tasks with length of 15 to 30 rounds. The models were evaluated on test data streams with the task lengths of 1-15, 20-35, 40-55 and 60-75 and the each stream consisted of 400 total tasks. For every training and test round, we used a mini-batch of 32 images for a single update. 
    
    In this online setup, each example occurs only once for a single task; and thus longer the length of a task means more data to adapt the models for that task. So the task lengths refer to different data regimes. 
    For instance, there are 32 to 480 examples for a task with length 1-15 whereas there are 1920 to 2400 examples when the task length ranges from 60-75.
    
    The test results are shown in Figure~\ref{fig:online_cifar}. The average task accuracy increases as there is more data available with the growing task length. The Sparse-MetaNet model outperforms the baselines across all the test streams and the performance gap is large for the stream consisting of tasks with lengths ranging from 1 to 15. In terms of the perseveration error, the fine-tuning with pretraining baseline achieves lower rates. However, the perseveration error rates tend to increase for the baseline approaches with more examples provided. In contrast, the Sparse-MetaNet results show decreasing error rates. Our Sparse-MetaNet also achieves the best interference error rate.
    
\subsection{Adaptive Language Modelling}
    Natural language text can provide a large scale testbed for sequential adaptation methods due distribution shifts over vocabulary and topics across text~\cite{blitzer2007domain}. We evaluated Transformer-XL networks~\cite{dai2019transformer} based on Sparse-MetaNets on character and word-level language modelling on enwik8~\cite{mahoney2011large} and WikiText-103~\cite{merity2016pointer} benchmarks. Particularly, we trained the 12L Transformer-XL model on enwik8 and the Transformer-XL Standard model on WikiText-103 data. Note that as those models consist of 12 and 16 transformer blocks and 8 and 10 attention heads per block (41 and 151 million parameters total), they impose significant challenges.
    
    We augmented all attention layers but the first layer of the FF component with the fast-weights. For enwik8, we apply the fast-weights to the embeddings layer as well. Since the Transformer-XL models have a large number of parameters and they are already computation intensive to train, we generated extremely sparse fast-weights with a sparsity of 0.95 at each time step by using a very small value for the mask probability parameter $p=0.05$. 
    
    The fast-weights were accumulated no more than $k$ steps and they were reset to zero after every BPTT step $k$, which we set to 5 for enwik8 and 4 for WikiText-103. Moreover, the fast-weights were shared for mini-batch examples that are computed on the same GPU and total 4 V100 GPUs were used to train each model.
    
    In this experiment, we adjusted the batch size to fit in the GPU memory and used the same hyper-parameters as the original Transformer-XL training. When testing the models, we increased the mask parameter $p$ to 0.5 for enwik8 and 0.3 for WikiText-103 based on the validation results. The fast-weights were gathered across time steps as described in Algorithm~\ref{alg:sparse_metanet} providing a long-term memory. After a ablation, we found that too small or too large $p$ hurts the performance and the fast-weight accumulation always improves the performance on the validation data.
    
    Our final results are reported in Table~\ref{tab:enwik8} and~\ref{tab:wt103} along with results from previous models with the similar number of parameters. The Sparse-MetaNet models obtained 1.00 bpc for character-level language modelling on the enwik8 and 22.67 perplexity for word-level language modelling on the WikiText-103 datasets by performing the online fast-weight updates. The performance gains over the offline Transformer-XL results are 0.06 bpc and 1.33 perplexity, respectively.
    
    \begin{table*}[t]
        \caption{
            Enwik8 test results. Total meta-weights: 63K.
        }
        \label{tab:enwik8}
        \small
        \centering
        \begin{tabular}{l|cl}
            \toprule
            \bf Model & \bf \#parameters & \bf bpc \\
            \midrule
                RHN \cite{zilly2017recurrent} & 46M & 1.27 \\
                mLSTM \cite{krause2016multiplicative} & 46M & 1.24 \\
                12L Transformer \cite{al2019character} & 44M & 1.11 \\
                Adaptive-Span \cite{sukhbaatar2019adaptive} & 39M & 1.02 \\
            \midrule
            12L Transformer-XL~\cite{dai2019transformer} & 41M & 1.06  \\
            12L Sparse-MetaTXL (This work) & 41M (+ 63K) & \textbf{1.00 (-0.06)} \\
            \bottomrule
        \end{tabular}

    \end{table*}
    
    \begin{table*}[t]
        \caption{
    		WikiText-103 test results. Total meta-weights: 78K
    	}
    	\label{tab:wt103}
    	\small
    	\centering
    	\begin{tabular}{l|cl}
    		\toprule
    		\bf Model & \bf \#parameters &  \bf PPL \\
    		\midrule
    		     LSTM \cite{grave2016improving} & - & 48.7 \\
        		 LSTM + Cache \cite{grave2016improving} & - & 40.8 \\
        		 QRNN \cite{merity2018analysis} & 151M & 33.0 \\
        		LSTM + Hebbian + Cache \cite{rae2018fast} & - & 29.9 \\
    		\midrule
    		16L Transformer-XL~\cite{dai2019transformer} & 151M & 24.0 \\
    		16L Sparse-MetaTXL (This work) & 151M (+ 78K) & \textbf{22.67 (-1.33)} \\
    		\bottomrule
    	\end{tabular}
    	
    \end{table*}

\section{Conclusion}
The only universal learning algorithm that we are aware of is how humans learn. Human learning is robust and flexible -- it relies on causality, has an ability of fast and sequential adaptation and balances memory encoding and active forgetting, across a large number of familiar and unfamiliar scenarios.
Meta-learning offers a promising computational paradigm to learn such a universal learning algorithm in a data-driven way.

In this work, we proposed a meta-learning approach to learn a sequential adaptation algorithm for arbitrary deep neural network architectures.
Our approach performs sequential adaptation with a bounded compute and memory across changing environment and tasks.
The proposed Online Cifar setup can serve as a useful benchmark for studying flexible models and algorithms that go beyond the fixed distribution regime. 

In the current state of the Sparse-MetaNet method, a sparsity mask is sampled from a fixed distribution. A future work should explore learning-based approaches for a conditional mask distribution, so that a Sparse-MetaNet model can selectively encode a fast-weight memory from past gradients. The current work has a limited focus on the catastrophic interference issue in neural networks. A future work can extend the Sparse-MetaNet approach for mitigating this issue.

\section*{Broader Impact}
The sequential adaptation approach presented in this work can enable deep neural network models that can be corrected (i.e. updated) on the fly. One can correct a certain bias of the model by providing feedback via input data stream once the bias is discovered. 

As learning algorithm itself is parameterized and learned from data, meta-learning approaches generally introduce another layer of difficulty in studying hidden biases. One needs to probe learned learning algorithm as well as its underlying prediction model. As we have done in our WCST experiment, adapting human cognitive and psychology tests could provide useful tools that can help understand behavior and reveal biases of learned algorithms and its underlying prediction models.

\small
\bibliography{example_paper}
\bibliographystyle{plain}

\clearpage

\appendix

\section{Training Details}
    For the meta-learners in our Sparse-MetaNet models, we used a 3-layer feed-forward neural net with 20 hidden units and LeakyReLU as non-linearity. Each fast-weight branch was optimized by a separate meta-learner. We did not use higher-order gradients as done in MAML~\cite{finn2017model}. The meta-learner inputs were pre-processed as in~\cite{andrychowicz2016learning}.
    
    For the WCST experiment, some Sparse-MetaNet runs computed zero activations during the forward passes due to unlucky initialization~\cite{frankle2018lottery} and the optimization did not advance at all. In this case, we restarted the runs for better random initialization to advance the training.
    
    The 12-layer ResNet architecture consists of 4 residual blocks with 64, 96, 128 and 256 filters. 
    For the online baselines and the Sparse-MetaNet model, the BatchNorm layer~\cite{ioffe2015batch} stats was reset at the beginning of an entire dev/test stream while for the offline baselines, it was reset for each task.
    
    We set the batch sizes to 24 and 72 for enwik8 and WikiText-103 experiments. During training, each fast-weight update was performed with 128 characters (50 words) for $k-1$ steps. Afterwards, 512 characters (150 words) were used for the gradient update of the slow and meta-weights at the $k$-th step. A full unsharing of the fast-weights by setting the batch size to 1 yielded the best result on validation data for enwiki8. 
    For WikiText-103, however, our best result was obtained by setting the batch size to 4.
    The Transformer-XL attention span was expanded for 4000 characters (700 words) for evaluation. Table~\ref{tab:hyperparam} lists the Sparse-MetaNet hyper-parameters.

    \begin{table*}[ht]
      \caption{Hyperparameters used in the experiments}
      \label{tab:hyperparam}
      \centering
        \small
      \begin{tabular}{l lllll llll} 
        \toprule
        
        {} & \multicolumn{4}{c}{Train}  &\multicolumn{5}{c}{Evaluation}\\
        \cmidrule(l{3pt}r{3pt}){2-6} \cmidrule(l{3pt}r{3pt}){7-10}
        
        & $k$ & $\gamma$ & $\beta_1$ & $\beta_2$ & $p$ & $\gamma$ & $\beta_1$ & $\beta_2$  & $p$ \\
        \hline
        WCST & 3 & 0.9 & 0.5 & 0.5 & 0.3 & 0.9 & 0.5 & 0.5 & 0.3 \\
        Online Cifar  & 3 & 0.99 & 0.5 & 0.5 & 0.3 & 0.99 & 0.5 & 0.5 & 0.5 \\
        Enwik8 & 5 & 0.0 & 0.0 & 1.0 & 0.05 & 0.999 & 0.5 & 0.5 & 0.5 \\
        WikiText-103 & 4 & 0.0 & 0.0 & 1.0 & 0.05 & 0.999 & 0.5 & 0.5 & 0.3 \\
        \bottomrule
      \end{tabular}
      \vspace{-5pt}
    \end{table*}

\section{WCST online agent results}
\begin{figure*}[!t]
        \begin{center}
        \includegraphics[width=1\textwidth]{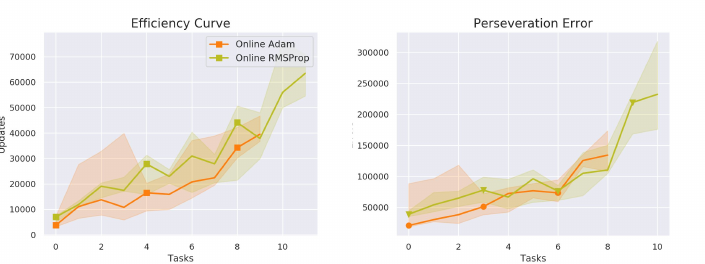}
        \caption{WCST online agent results. Left: the number of updates required to solve each task. Right: total perseveration errors made by each agent before switching to the next task. The online agents diverge as they see more tasks and fail to converge after solving around 10 tasks.}
        \label{fig:wcst_offline}
        \end{center}
        \vskip -0.2in
    \end{figure*}

    In Figure~\ref{fig:wcst_offline}, we showed the results of the online agents on the WCST task. The online agents diverge as they see more tasks and fail to converge after solving around 10 tasks

\newpage
\section{Code Snippet for WCST Task Generation}
\tiny
\begin{lstlisting}
    import random
    import numpy as np
    
    
    class WCST(object):
    	"""WCST Generator"""
    	def __init__(self):
    		super(WCST, self).__init__()
    		self.inp_codes = np.array([[0,0,0,1], [0,0,1,0], [0,1,0,0], [1,0,0,0]], dtype=np.float32)
    		self.task_dict = {'color':0, 'shape':1, 'number':2}
    		
    		self.current_task = None
    		self.old_task = None
    
    		self.current_trial_task = 0
    		self.total_trial = 0
    		self.total_task = 0
    		self.tasks_taken = []
    		self.tasks_scheduled = None
    
    	def next_task(self, allow_same_task=False):
    		if self.tasks_scheduled is not None:
    			self.current_task = self.tasks_scheduled[self.total_task]
    		else:
    			self.old_task = self.current_task
    			self.current_task = random.randint(0, 2)
    			while self.current_task == self.old_task and not allow_same_task:
    				self.current_task = random.randint(0, 2)
    
    		self.current_trial_task = 0
    		self.total_task += 1
    		self.tasks_taken.append(self.current_task)
    
    	def next_card(self, n=1):
    		inps_code = []
    		targets = []
    		targets_perv = []
    		for i in range(n):
    			# sample color, shape and number codes in that order
    			code_inds = random.sample(range(4), 3)
    			target = code_inds[self.current_task]
    			if self.old_task is not None:
    				target_perv = code_inds[self.old_task]
    			else:
    				target_perv = None
    
    			inp_code = self.inp_codes[code_inds].reshape(1, -1)
    
    			self.current_trial_task += 1
    			self.total_trial += 1
    
    			if n == 1:
    				return inp_code, target, target_perv
    
    			inps_code.append(inp_code)
    			targets.append(target)
    			targets_perv.append(target_perv)
    
    		inps_code = np.concatenate(inps_code, axis=0)
    
    		return inps_code, targets, targets_perv

\end{lstlisting}

\newpage
\section{Code Snippet for Online Cifar Task Generation}
\tiny
\begin{lstlisting}
    
    import sys
    import numpy as np
    import random
    import pickle
    
    
    class Task(object):
    
    	def __init__(self, task_map, new_old_map=dict(), kept_ids=list()):
    		self.task_map = task_map
    		self.new_old_map = new_old_map
    		self.kept_ids = set(kept_ids)
    
    
    class OnlineCifarGenerator(object):
    	"""Online Cifar Generator"""
    	def __init__(self, data_dir='path_to/cifar10/cifar-10-batches-py',
    				 nb_classes=5, nb_classes_perv=2, nb_classes_kept=1, nb_samples_per_class=100000000, 
    				 batchsize=32, split='train'):
    		super(OnlineCifarGenerator, self).__init__()
    		self.data_dir = data_dir
    		self.nb_classes = nb_classes
    		self.nb_samples_per_class = nb_samples_per_class
    		self.batchsize = batchsize
    		self.split = split
    
    
    		self.nb_classes_perv = nb_classes_perv
    		self.nb_classes_kept = nb_classes_kept
    		
    		self._load_data(self.data_dir)
    
    		self.current_task = None
    		self.old_task = None
    
    		self.total_iter = 0
    		self.total_task = 0
    		self.current_task_epoch = 0
    		self.current_task_iter = 0
    
    	def _load_data(self, data_dir):
    		if self.split == 'train':
    			data_path = data_dir + 'cifar-100-python/train.pk'
    		elif self.split == 'valid':
    			data_path = data_dir + 'cifar-100-python/valid.pk'
    		elif self.split == 'test':
    			data_path = data_dir + 'cifar-100-python/test.pk'
    		else:
    			raise ValueError("Incorrect value for split argument")
    
    		with open(data_path, 'rb') as fo:
    			self.data_dict_train = pickle.load(fo)
    
    		data_dict = {}
    		with open(data_dir + 'cifar-100-python/test', 'rb') as fo:
    			imagedict = pickle.load(fo)
    			for y, x in zip(imagedict[b'fine_labels'], imagedict[b'data']):
    				if y in data_dict:
    					l = data_dict[y]
    					l += [x.reshape(1, -1)]
    					data_dict[y] = l
    				else:
    					data_dict[y] = [x.reshape(1, -1)]
    
    		self.data_dict_test = data_dict
    
    	def next_task(self, next_task=None):
    		kept_ids = []
    		kept_lbls = []
    		perv_ids = []
    		perv_lbls = []
    		new_ids = []
    		new_lbls = []
    
    		all_ids = set(range(self.nb_classes))
    		all_lbls = set(self.data_dict_train.keys())
    		if next_task is None:
    			perv_old_id = None
    			if self.current_task is not None:
    				if self.nb_classes_kept > 0:
    					kept_ids = random.sample(self.current_task.task_map.keys(), self.nb_classes_kept)
    					kept_lbls = [self.current_task.task_map[idc] for idc in kept_ids]
    					all_ids = all_ids - set(kept_ids)
    					all_lbls = all_lbls - set(kept_lbls)
    				if self.nb_classes_perv > 0:
    					perv_old_id = random.sample(all_ids, self.nb_classes_perv)
    					perv_lbls = [self.current_task.task_map[idc] for idc in perv_old_id]
    					perv_ids = random.sample(all_ids, self.nb_classes_perv)
    					all_ids = all_ids - set(perv_ids)
    					all_lbls = all_lbls - set(perv_lbls)
    				new_ids = random.sample(all_ids, self.nb_classes-self.nb_classes_kept-self.nb_classes_perv)
    				new_lbls = random.sample(all_lbls, self.nb_classes-self.nb_classes_kept-self.nb_classes_perv)
    
    				task_ids = kept_ids + perv_ids + new_ids
    				task_lbls = kept_lbls + perv_lbls + new_lbls
    				task_map = dict(zip(task_ids, task_lbls))
    
    				new_old_map = dict()
    				if perv_old_id is not None:
    					new_old_map = dict(zip(perv_ids, perv_old_id))
    				current_task = Task(task_map, new_old_map, kept_ids)
    			else:
    				task_ids = random.sample(all_ids, self.nb_classes)
    				task_lbls = random.sample(all_lbls, self.nb_classes)
    				task_map = dict(zip(task_ids, task_lbls))
    				current_task = Task(task_map)
    		else:
    			current_task = next_task
    		self.old_task = self.current_task
    		self.current_task = current_task
    
    		train_x = []
    		train_y = []
    		train_perv_y = []
    		train_kept_y = []
    
    		test_x = []
    		test_y = []
    		test_perv_y = []
    		test_kept_y = []
    		for idc, lbl in self.current_task.task_map.items():
    			x = self.data_dict_train[lbl]
    			x = random.sample(x, min(len(x), self.nb_samples_per_class))
    			n_x = len(x)
    			train_x.extend(x)
    			train_y.extend([idc]*n_x)
    			old_idc = self.current_task.new_old_map.get(idc, -1)
    			train_perv_y.extend([old_idc]*n_x)
    			is_kept = idc in self.current_task.kept_ids
    			train_kept_y.extend([is_kept]*n_x)
    
    			x = self.data_dict_test[lbl]
    			n_x = len(x)
    			test_x.extend(x)
    			test_y.extend([idc]*n_x)
    			test_perv_y.extend([old_idc]*n_x)
    			test_kept_y.extend([is_kept]*n_x)
    
    		self.train_x = np.array(train_x, dtype=np.float32).squeeze()
    		self.train_y = np.array(train_y, dtype=np.int32)
    		self.train_perv_y = np.array(train_perv_y, dtype=np.int32)
    		self.train_kept_y = np.array(train_kept_y)
    
    		self.test_x = np.array(test_x, dtype=np.float32).squeeze()
    		self.test_y = np.array(test_y, dtype=np.int32)
    		self.test_perv_y = np.array(test_perv_y, dtype=np.int32)
    		self.test_kept_y = np.array(test_kept_y)
    		
    		self.n_train = self.train_x.shape[0]
    		self.n_test = self.test_x.shape[0]
    
    		self.train_idx = range(self.n_train)
    		self.test_idx = range(self.n_test)
    		np.random.shuffle(self.train_idx)
    		np.random.shuffle(self.test_idx)
    
    		self.total_task += 1
    		self.current_task_epoch = 0
    		self.current_task_epoch_iter_train = 0
    		self.current_task_epoch_iter_test = 0
    		self.current_task_iter = 0
    
    	def next(self, train=True):
    		epoch_done = False
    		if train:
    			idx = self.train_idx[self.current_task_epoch_iter_train*
    			self.batchsize:(self.current_task_epoch_iter_train + 1)*self.batchsize]
    			
    			x = self.train_x[idx]
    			y = self.train_y[idx]
    			perv_y = self.train_perv_y[idx]
    			kept_y = self.train_kept_y[idx]
    
    			self.current_task_epoch_iter_train += 1
    			self.current_task_iter += 1
    			self.total_iter += 1
    
    			if self.current_task_epoch_iter_train*self.batchsize > self.n_train:
    				np.random.shuffle(self.train_idx)
    				self.current_task_epoch += 1
    				self.current_task_epoch_iter_train = 0
    				epoch_done = True
    
    		else:
    			idx = self.test_idx[self.current_task_epoch_iter_test*
    			self.batchsize:(self.current_task_epoch_iter_test + 1)*self.batchsize]
    			
    			x = self.test_x[idx]
    			y = self.test_y[idx]
    			perv_y = self.test_perv_y[idx]
    			kept_y = self.test_kept_y[idx]
    
    			self.current_task_epoch_iter_test += 1
    
    			if self.current_task_epoch_iter_test*self.batchsize > self.n_test:
    				self.current_task_epoch_iter_test = 0
    				epoch_done = True
    
    		return x, y, perv_y, kept_y, epoch_done

\end{lstlisting}
\end{document}